\documentclass[10pt,twocolumn,letterpaper]{article}

\usepackage{iccv}
\usepackage{times}
\usepackage{epsfig}
\usepackage{graphicx}
\usepackage{amsmath}
\usepackage{amssymb}
\usepackage{multirow}
\usepackage{bm}
\usepackage{booktabs}

\usepackage[pagebackref=true,breaklinks=true,letterpaper=true,colorlinks,bookmarks=false]{hyperref}

\iccvfinalcopy 

\def\iccvPaperID{6} 

\ificcvfinal\fi

\begin{document}

\title{Improved Positional Encoding for Implicit Neural Representation based Compact Data Representation}

\author{Bharath Bhushan Damodaran, Francois Schnitzler, Anne Lambert, Pierre Hellier \\
InterDigital, Inc.,\\
Rennes, France\\
{\tt\small bharath.damodaran@interdigital.fr}
}

\maketitle
\ificcvfinal\thispagestyle{empty}\fi

\begin{abstract}
Positional encodings are employed to capture the high frequency information of the encoded signals in implicit neural representation (INR). In this paper, we propose a novel positional encoding method which improves the reconstruction quality of the INR. The proposed embedding method is more advantageous for the compact data representation because it has a greater number of frequency basis than the existing methods. Our experiments shows that the proposed method achieves significant gain in the rate-distortion performance without introducing any additional complexity in the compression task and higher reconstruction quality in novel view synthesis.    

\end{abstract}

\section{Introduction}\label{sec:intro}
Implicit neural representation (INR) or neural radiance field (NRF) has gained popularity recently, due to its capability  to represents different kinds of multi-dimensional signals~\cite{mildenhall2020nerf}. INR represents a signal by over-fitting a continuous function (neural network) which takes as input the coordinates of the signal and outputs the pixel color values~\cite{tancik2020fourier, sitzmann2020implicit}, in the case of signal regression. By over-fitting, INR learns to compactly represent the signals in the lower-dimensional space by discarding irrelevant information when the number of parameters is lower than the length of the underlying signal. The representation capacity of the INR depends on the number of parameters used to approximate the signals. Thus, the approximation quality can be adjusted by changing the architecture of the INR network. INR can be used for compact data representation~\cite{pmlr-v162-dupont22a,bauer2023spatial},  image and video compression, since storing an image/video amounts to storing the weights for the neural network~\cite{dupont2021coin,schwarz2022metalearning, chen2021nerv, damodaran2023dcc,schwarz2023modality}. Reconstructing the image or signal then amounts to extracting the weights and evaluating the neural network for all coordinates.

Directly using a multi-layer perceptron (MLP) results in an overly smooth reconstruction due to the spectral bias of the regular MLP. Tancik~\etal\cite{tancik2020fourier} showed MLP with ReLU activation functions are not suitable to encode signals with high frequency content. To overcome the spectral bias, Sitzmann~\etal\cite{sitzmann2020implicit} replaced ReLU with sine activation function whereas Tancik~\etal\cite{tancik2020fourier} used positional encoding with random Fourier features followed by MLP with ReLU activation function. 


Positional encoding using random Fourier features maps the input coordinates to a high dimensional embedding as $\gamma(\mathbf{x}):\mathcal{R}^d \rightarrow \mathcal{R}^{2D}$ using $D$ dimensional random basis. The embedding dimension as well as the number of random Fourier basis has a direct impact on the reconstruction quality. However, in Tancik~\etal\cite{tancik2020fourier}, the number of random Fourier basis used is only half the size ($D$) of the embedding dimension ($2D$). Thus, the existing Fourier embedding used in the INR is limited in covering the frequency spectrum and also in its reconstruction quality, especially when the embedding dimension is small.

In this paper, we propose an alternative positional embedding method for INR to improve reconstruction quality. Our proposed embedding method contains the same number of basis vectors as the embedding dimension. Therefore, it covers the frequency spectrum better. 
Additionally, our proposed method does not increase the encoding or decoding complexity of the signals nor the size of the bit-stream in compression tasks. We evaluated our proposed method on image reconstruction task, image compression and novel view synthesis and the results showed that our proposed method has significant gains 
without any additional increase in bit-stream size and complexity.

The rest of the paper is organized as follows: Section~\ref{sec:INR} introduces  the implicit neural representation and existing Fourier feature mapping, section~\ref{sec:proposed} presents our proposed method, section~\ref{sec:exp} reports on our experimental results, and conclusion is derived in section~\ref{sec:con}.

\section{Implicit Neural Representation}\label{sec:INR}
Let $\mathbf{I} \in \mathbb{R}^{W \times H \times 3}$ be a color image, $x,y \in \mathbb{R}$ be the pixel coordinates in the normalized range $[-1, 1]$, $I(x,y)$ denotes the pixel values (RGB) at the coordinates $x,y$ and $\gamma(x,y)$ a positional encoding of the coordinates. The INR is a neural network $f_{\bm{\theta}}$, parameterized by the weights $\bm{\theta}$ such that it maps the given coordinates to the pixel intensity values (RGB). In other words, $\forall x,y, f_{\bm{\theta}}(\gamma(x,y)) \sim I(x,y)$. Without loss of generality, it can be extended to any multi-dimensional signals. 

The weights $\bm{\theta}$ of the INR are estimated by over-fitting (minimizing) the following loss function
\begin{equation}\label{eq:inr_mse}
\bm{\theta}^* =\arg \min_{\bm{\theta}} L(x,y, \bm{\theta})=\frac{1}{N} \sum_{x,y} d(I(x,y),f_\mathbf{\bm{\theta}} (\gamma(x,y))),
\end{equation}
where the sum is over all the pixels in the image $(N =W\times H)$, $W, H$ is width and height of the image, $d$ is any distortion metric which measures the discrepancy between the predicted (reconstructed) pixels by $f_{\bm{\theta}}$ and the actual pixel values of the image $I$. The metric $d$ is preferably a differentiable distortion measure, such as mean squared error or perceptual metric such as LPIPS. In this paper, mean squared error (MSE) is used as the distortion metric. Once the equation \eqref{eq:inr_mse} is optimized, at the inference image can be reconstructed by evaluating $f_{\bm{\theta^*}}$ over all the pixel co-ordinates.

\begin{figure}
    \centering
    \includegraphics[scale=0.4]{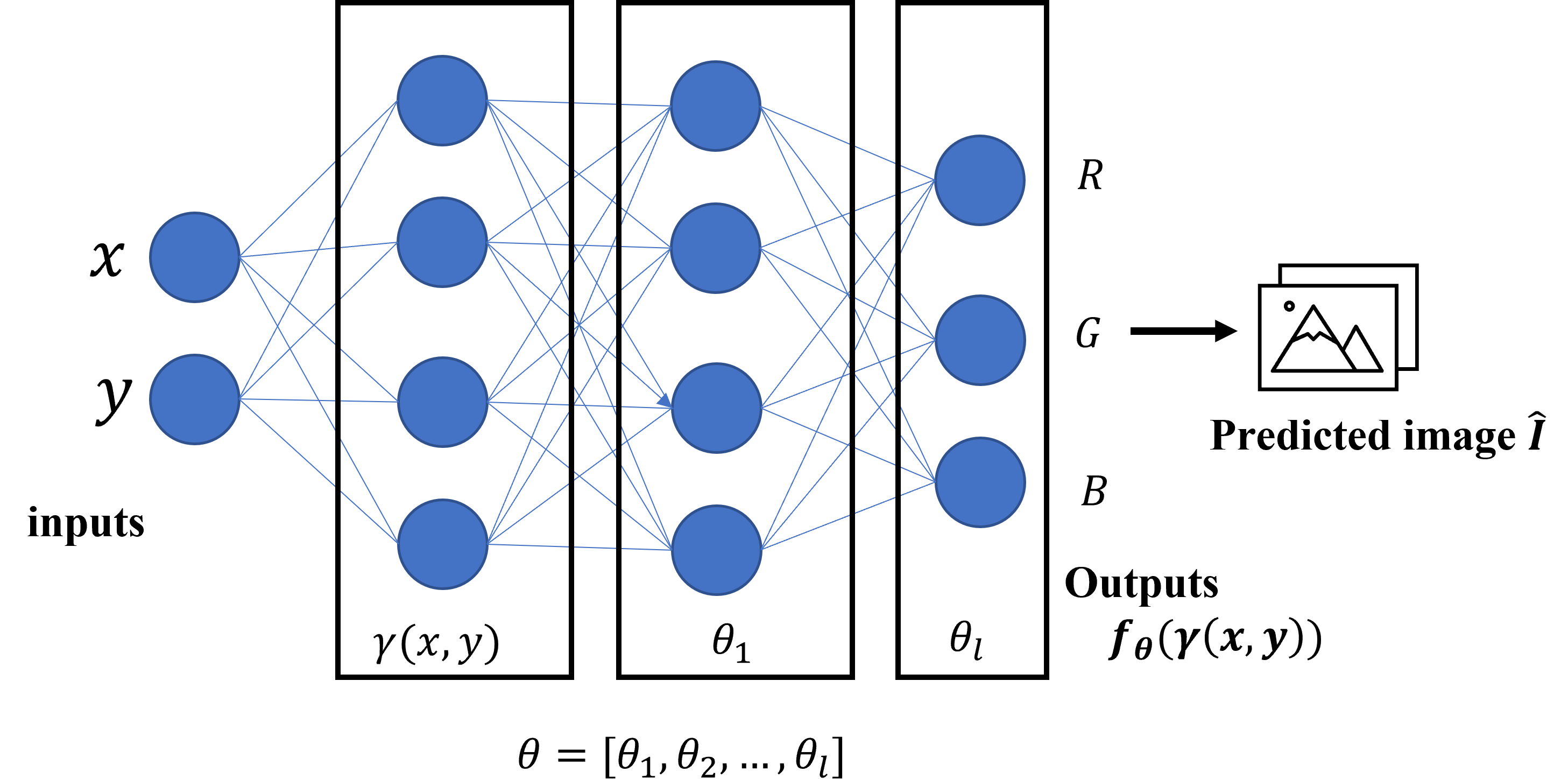}
    \caption{Implicit neural representation (INR) with Fourier feature mapping. $x,y\in\mathbb{R}$ is the normalized input coordinates, $\gamma(x,y)$ is the Fourier feature mapping, $\bm{\theta}'s$ are weights of the MLP. For the compression, $\bm{\theta}$'s are encoded in the bit-stream and transmitted to the decoder side.}
    \label{fig:INR}
\end{figure}

\textbf{Data representation:} The optimized $\bm{\theta}^*$ can be used as the data representation for the down-stream tasks. Compressing  an image $\mathbf{I}$ is equivalent to encoding the values of the weights $\bm{\theta}^*$ in the bit-stream. 
For compression it is not possible to choose large  neural network for better reconstruction quality as it would increase bit-length. Thus, the number of weights is constrained at the expense of the distortion. For each image $I$, there is one specific INR $f_\theta$ which is overfitted to the given image I. This is different from the end-to-end compression method~\cite{shukor2022video,balcilar2022reducing,balcilar2023latentshift}. To generate bit-streams of different sizes, INR is trained with different number of hidden layers and nodes. Figure \ref{fig:INR} illustrates an implicit neural network (INR) based image compression system.

\textbf{Positional Encoding with Fourier features:}
The Fourier feature mapping is based on the Bochner's theorem to approximate a shift-invariant kernels.
Tancik~\etal\cite{tancik2020fourier} used random Fourier feature (RFF) mapping $\gamma: \mathcal{R}^d \rightarrow \mathcal{R}^{2D}$ to the input coordinates before feeding them to MLP with ReLU activation functions. The Fourier feature mapping of the  coordinate $\bm{v}=\left(x,y\right)$ is defined as follows:
\begin{multline}
 \gamma\left(\bm{v}\right) \in \mathcal{R}^{2D}=[\cos{\left(2\pi \bm{w}_1^T\bm{v}\right),\sin{\left(2\pi \bm{w}_1^T\bm{v}\right),\ \ }\ldots,} \\  \cos{\left(2\pi \bm{w}_D^T\bm{v}\right),\sin{\left(2\pi \bm{w}_D^T\bm{v}\right)}}]^T,
 \label{eq:sin_cos_emb}
 \end{multline}
where the coefficients $\bm{w}_i$ are the Fourier basis frequencies when the mapping is seen as a Fourier approximation of a shift-invariant kernel function. The basis vectors $\bm{w}_i$'s are randomly sampled from the Gaussian distribution with appropriate band-width $\sigma$, i.e $\bm{w}_i \sim N(0, \sigma), i=1 \ldots D$. This positional encoding is used in the existing INR and NeRF literature.   

For the mapping dimension of $2D$, only $D$ number of Fourier basis are sampled. 
One could note that the number of sampled frequencies is half the number of the mapping size. This could be an issue when the mapping size is small, which is the case in compression tasks. 

\section{Proposed method}\label{sec:proposed}
To increase the number of random Fourier basis in the Fourier feature mapping, here we propose an alternative positional encoding which is also based on the Bochner's theorem, and we label our proposed method as \emph{RFF-cosine mapping}. 
Let $\phi:\mathcal{R}^d \rightarrow \mathcal{R}^{2D}$ be the mapping of the our proposed RFF-cosine mapping, which maps the input coordinates to the RFF-cosine feature mapping as
\begin{multline}
\phi\left(\bm{v}\right)=[\sqrt2\cos\left(2\pi \bm{w}_1^T\bm{v}+b_1\right),\sqrt2\cos\left(2\pi \bm{w}_2^T\bm{v}+b_2\right),\ \   \\ \ldots, 
\sqrt2\cos{\left(2\pi \bm{w}_{2D}^T\bm{v}+b_{2D}\right)}]^T ,      
\label{eq:cos_emb}
\end{multline}
where $\bm{w}_i’s, i=1, \ldots, 2D$ are randomly sampled from the Gaussian distribution with the bandwidth parameter $\left(\sigma\right)$, and bias vectors $b_i's$ are randomly sampled from the uniform distribution in $\left[0,2\pi\right]$. 
The advantage of using our proposed method is directly evident by comparing eqn \eqref{eq:cos_emb} and \eqref{eq:sin_cos_emb}, where our proposed RFF-cosine feature mapping has more (twice) number Fourier basis frequencies with the same mapping size as compared to eqn~\eqref{eq:sin_cos_emb}. This allows to encode high frequency information in the signals better than the existing  positional encoding method. By replacing our proposed RFF-cosine mapping in eqn~\eqref{eq:inr_mse}, the loss function to be minimized is as follows
\begin{equation}\label{eq:inr_mse_rffcosine}
\bm{\theta}^* =\arg \min_{\bm{\theta}} L(x,y, \bm{\theta})=\frac{1}{N} \sum_{x,y} d(I(x,y),f_\mathbf{\bm{\theta}} (\phi(x,y))).
\end{equation}
%
%
\begin{table*}[!t]
    \centering
    \caption{Rate distortion (RD) performance of our proposed embedding method and the existing one for different bit-rates with different mapping sizes evaluated on the subset (images 5 to 14) of Kodak dataset . Q1-Q4 are different MLP architectures. Rates are measured in bits per pixel (bpp), and distortion in PSNR. The best results are in \textbf{bold}. }
    \vspace{2mm}
    \label{tab:RD}
    \begin{tabular}{l|l|l|l|l|l|l}
    \toprule
      Mapping size& \multicolumn{2}{c}{Method}  & Q1 & Q2 & Q3 & Q4  \\ \midrule
        %
               \multirow{4}{*}{8} & \multicolumn{2}{c|}{BPP}  &   0.0782  &  0.1661 &  0.3111 &  0.6202    \\ \cmidrule{2-7} 
         &\multirow{2}*{PSNR}& Existing method & 18.26  & 18.70& 19.04   & 19.38 \\    
        &  &  Ours &  \textbf{20.33}  & \textbf{20.83} &  \textbf{21.37} &  \textbf{22.01}      \\ \midrule
         \multirow{4}{*}{16} & \multicolumn{2}{c|}{BPP}  &   0.0848  &  0.1759 & 0.3202 & 0.633    \vspace{0.5mm}\\ \cmidrule{2-7} 
         &\multirow{2}*{PSNR}& Existing method & 21.67  & 22.48&22.69   & \textbf{23.51} \\    
         &  & Ours & \textbf{22.15}  &  \textbf{22.79} &\textbf{22.85} & {23.50}      \\ \midrule
        \multirow{4}{*}{32} & \multicolumn{2}{c|}{BPP}   & 0.0977  & 0.1954  & 0.3385 & 0.6593      \\ \cline{2-7} 
         &\multirow{2}*{PSNR}& Existing method &  22.42  & 22.94& 23.16   & 23.82   \\ 
        & & Ours & \textbf{22.56}   & \textbf{23.11} &\textbf{23.38} & \textbf{23.95}   \\ \midrule
        \multirow{4}{*}{64} &\multicolumn{2}{c|}{BPP} &  0.1238 & 0.2345 & 0.3750 & 0.7114     \\ \cmidrule{2-7}
         &\multirow{2}*{PSNR}& Existing method &  22.96 & 23.40 & 23.75   & 24.19   \\
        & & Ours & \textbf{23.14}  & \textbf{23.45} &\textbf{23.79} & \textbf{24.33}    \\ \bottomrule
    \end{tabular}
\end{table*}
For the signal compression, as the Fourier embeddings are generated from randomly sampled frequency basis, it is not necessary to encode the frequency basis in the bitstream, we only  need to write the random seed used to sample frequency basis in the bitstream. Our proposed method needs only to write one additional random seed used for the sampling from the uniform distribution, if different seeds are used.  Thus, our proposed positional encoding increases neither the size of the bitstream nor the complexity of encoding or decoding the signals.

%
\begin{figure}
    \centering
    \includegraphics[scale=0.45]{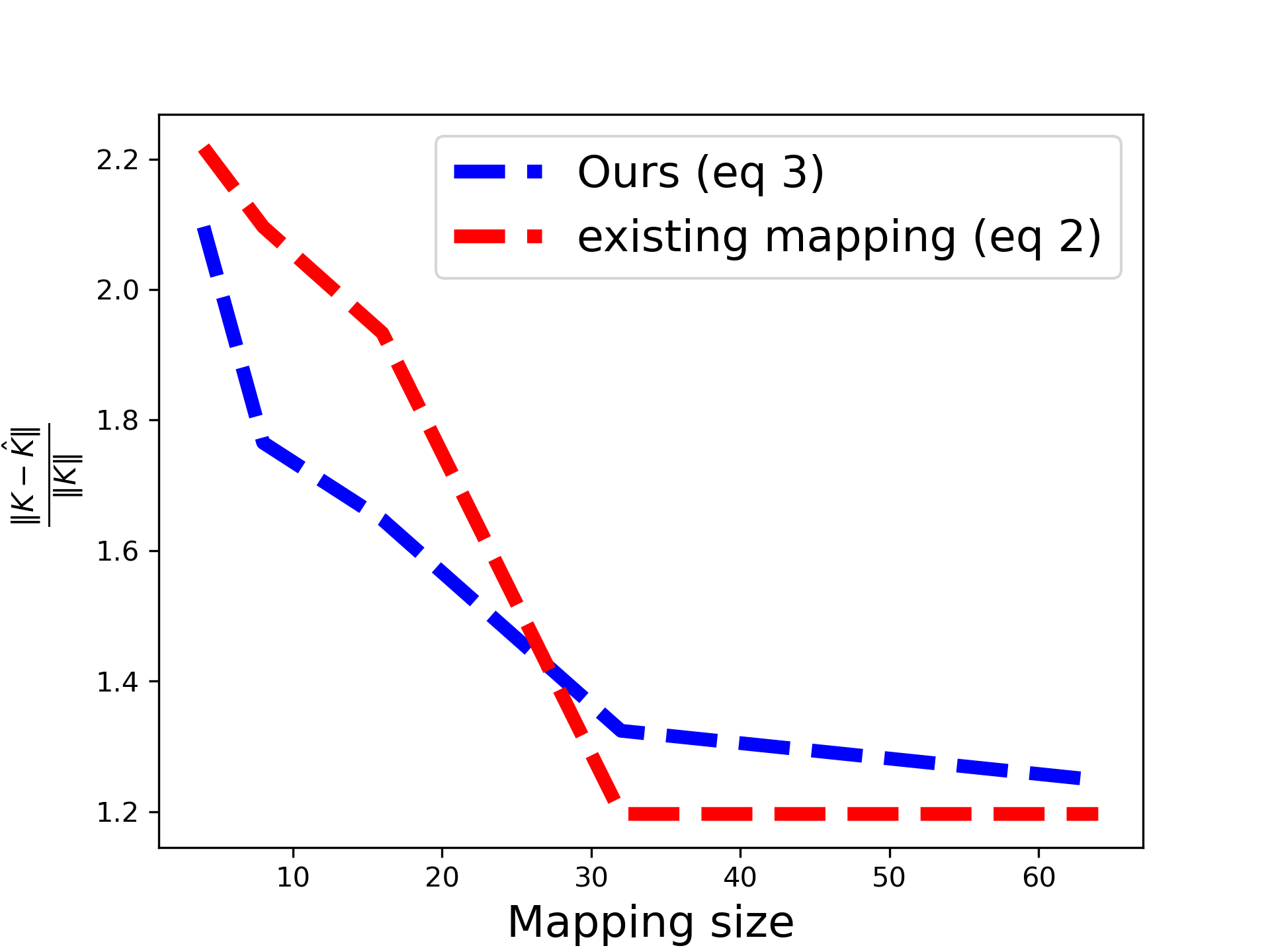}
    \caption{Kernel approximation error $\frac{\Vert \bm{K-\hat{K}}\Vert}{\Vert \bm{K}\Vert}$ of the $1D$ sine wave function $\left(3sin(t)+U(0,1), t\in[0,8\pi]\right)$ using our proposed mapping and existing mapping. $\bm{K}$ is the Gaussian RBF kernel matrix (with std dev. $\sigma$ computed using $5^{th}$ percentile of pairwise distances), and $\bm{\hat{K}}$ is the approximated kernel matrix using either our mapping~\eqref{eq:cos_emb} or existing mapping~\eqref{eq:sin_cos_emb}. Lower approximation error is better.}
    \label{fig:kernel_approx}
\end{figure}

\textbf{Illustration:} The benefit of our proposed feature mapping can be analyzed through the lens of neural tangent kernel (NTK)  and kernel approximation. Due to the greater number of frequency basis, the eigenvalue decay of the NTK of our proposed feature mapping might be slower than the existing Fourier feature mapping in~\eqref{eq:sin_cos_emb}, thus our method can encode more frequency content of the signal in the low mapping size. However, in the higher mapping size the eqn~\eqref{eq:sin_cos_emb} might be able to cover the entire frequency spectrum of the signal, thus performs equally similar to our proposed mapping or better. This can be validated with the toy experiment to approximate the Gaussian kernel using both the feature mapping. Figure~\ref{fig:kernel_approx} reports the kernel approximation error of the $1D$ sine wave function (with $1000$ samples), and show that our proposed feature mapping has smaller approximation error in the lower the mapping size, and as the mapping size increases 
the existing mapping has better approximation error. Thus, validating our proposal to use our proposed feature mapping in the tasks where the large mapping size or large neural networks cannot be used. 

\section{Experimental Results}\label{sec:exp}
We evaluated our proposed RFF-cosine mapping method on different sets of tasks: 
image compression, and novel view synthesis. Further, we evaluated the impact of the mapping size, and showed the superiority of our proposed method in low mapping size settings.


\textbf{Image compression:} We choose a subset (images $5$ to 14) of the Kodak dataset~\cite{eastman_kodak_kodak_nodate}. The RGB PSNR and bits per pixel (BPP) are used as the distortion and rate measure. We use the MLP architecture with varying number of hidden layers and hidden nodes $\{(5,20),(5,30), (10, 30), (10, 40) \}$, for different bit-rates as in the \cite{dupont2021coin}. The loss function is minimized with Adam optimizer using learning rate $2e-4$ for $50K$ iterations. To choose bandwidth parameter $\sigma$ for the Fourier mapping, we performed experiments using few images with $\{1, 5, 10\}$ and we observed that $\sigma=1$ gives optimal performance for the compression task.
For the compression, we encode the weights and bias in the half-precision as in \cite{dupont2021coin}.

The rate-distortion performance of our proposed embedding method and the existing one is reported in Table~\ref{tab:RD} with different network architectures and mapping size. The results reveal that our proposed mapping outperforms the existing one across different bit-rates and with different mapping size. More specifically, our method has a  significantly better PSNR than the existing method in the low mapping size (more than 2dB PSNR in average) though the mapping size is the same, which proves the advantage of having more frequency basis.

\begin{figure}
    \centering
    \includegraphics[scale=0.6]{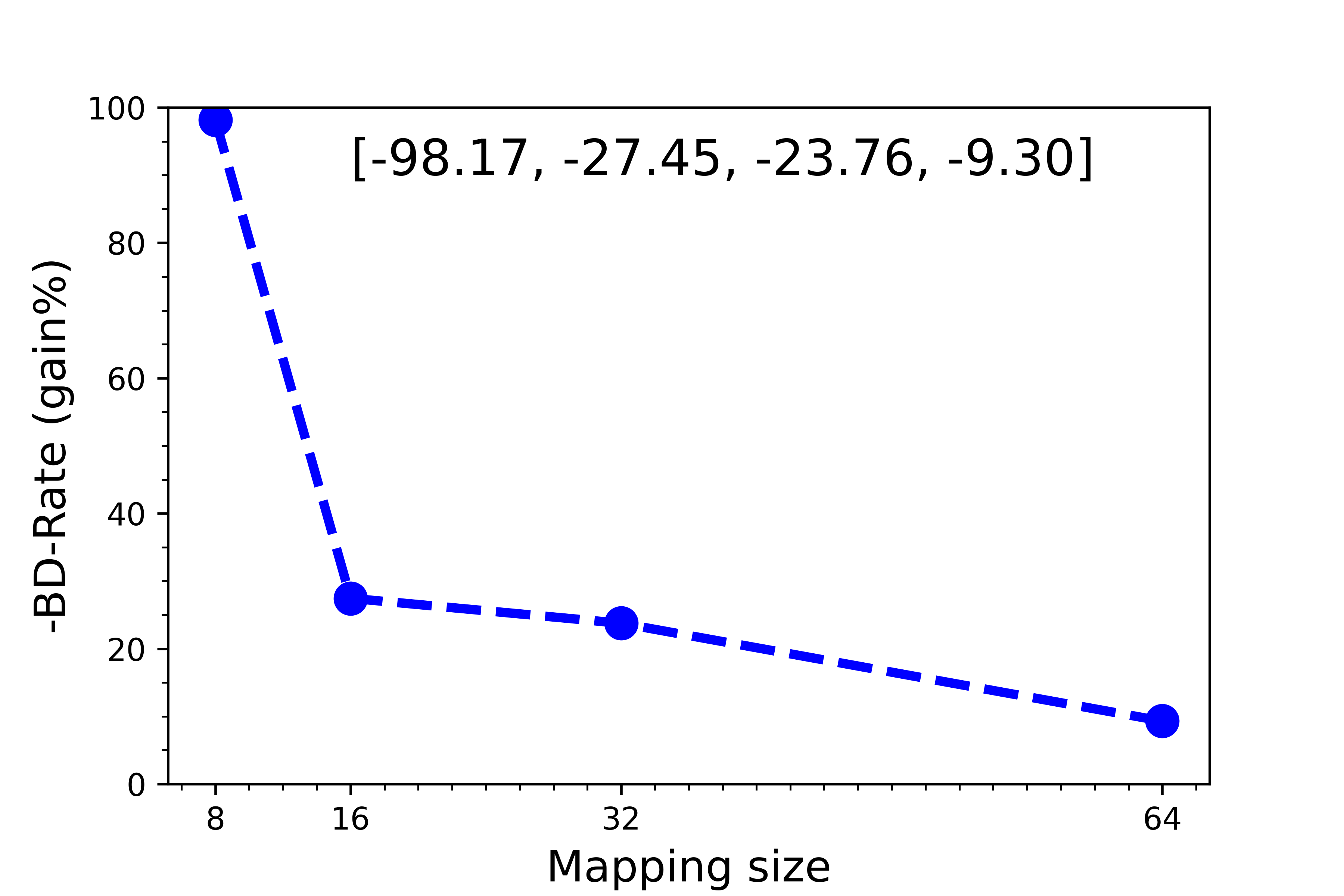}
    \caption{BD rate gain (in -\%) of our proposed method with respect to the existing method over various mapping size. The exact gains are displayed in text.}
    \label{fig:bdrate_ms}
\end{figure}

Further, to quantify the bitrate gain in $\%$ we  computed Bjontegard BD rate gain~\cite{bjontegaard2001calculation} and it is presented in figure~\ref{fig:bdrate_ms}. It demonstrates that our proposed embedding has about $98\%$ BD-rate gain at the low mapping size and about $10\%$ BD-rate gain at the higher mapping size. These gains are very significant in the compression literature, and they are achieved without any additional (encoding and decoding) complexity and increase in the size of bit-stream. 

\textbf{Novel view synthesis:} Finally, we evaluate our proposed embedding method on the novel view synthesis. For this, we applied our proposed method on the recent NeRF method named \emph{Nope-NeRF}~\cite{bian2023nope}, and compared with the existing Fourier features. We conducted experiments on the \emph{Ignatius} sequence from \emph{tanks} dataset, and after training we used 8 scenes to evaluate novel view synthesis quality. We evaluated the quality of the synthesis using PSNR, Structural Similarity Index Measure (SSIM) and Learned Perceptual Image Patch Similarity (LPIPS). We followed a similar experimental protocol to ~\cite{bian2023nope}. For the Fourier feature mapping, we conducted experiments with bandwidth parameter $\{1, 5, 10\}$ with mapping size of $60$  and we observed that $\sigma=1$ gave the best results.

Table~\ref{tab:nope-nerf} reports the reconstruction quality of the Nope-Nerf with our proposed positional encoding and existing encoding method, and showed that with our proposed method Nope-Nerf achieved best results.
\begin{table}[!htpb]
    \centering
    \caption{Reconstruction quality of the Nope-Nerf method for novel view synthesis with our proposed positional encoding and existing encoding. The best results are in \textbf{bold}. }
        \vspace{2mm}
    \begin{tabular}{c|c|c|c}
    \toprule
         Method & PSNR ($\uparrow$) & SSIM ($\uparrow$) & LPIPS ($\downarrow$) \\ \midrule
         Nope-NeRF+\eqref{eq:sin_cos_emb} & 23.79 & 0.60 & 0.50\\
          Nope-NeRF+Ours & \textbf{23.91} &\textbf{ 0.61} & \textbf{0.49} \\
         \bottomrule
    \end{tabular}

    \label{tab:nope-nerf}
\end{table}

\section{Conclusion}\label{sec:con}
In this paper, we proposed an improved positional encoding for implicit neural representation. The proposed embedding method has a greater number of Fourier frequency basis than the existing Fourier feature mapping used in the INR literature. The superiority of our method is evaluated on image compression and novel view synthesis tasks, and showed that it offered significant BD rate gain in compression, and better reconstruction quality in view synthesis. In the future work, we will explore the quantization aware training~\cite{damodaran2023dcc} and entropic coding to further improve the compression efficiency.  

%
{\small
\bibliographystyle{ieee_fullname}
\bibliography{egbib}
}
\end{document}